\documentclass[final]{cvpr}

\usepackage{times}
\usepackage{epsfig}
\usepackage{graphicx}
\usepackage{amsmath}
\usepackage{amssymb}

\usepackage[pagebackref=true,breaklinks=true,colorlinks,bookmarks=false]{hyperref}

\def\cvprPaperID{****} % *** Enter the CVPR Paper ID here
\def\confYear{CVPR 2021}

\begin{document}

%%%%%%%%% TITLE
\title{Gaze Estimation with an Ensemble of Four Architectures}
\author{Xin Cai$^{1}$, Boyu Chen$^{2}$, Jiabei Zeng$^{1}$, Jiajun Zhang$^{2}$, Yunjia Sun$^{1}$, \\
Xiao Wang$^{2}$, Zhilong Ji$^{2}$, Xiao Liu$^{2}$, Xilin Chen$^{1}$, Shiguang Shan$^{1}$\\
$^1$ Institute of Computing Technology, Chinese Academy of Sciences, Beijing 100190, China \\
$^2$ Tomorrow Advancing Life (TAL) Education Group \\
\{caixin20s, jiabei.zeng, sunyunjia18z, xlchen, sgshan\}@ict.ac.cn\\
\{chenboyu,zhangjiajun1, wangxiao15, jizhilong, liuxiao15\}@tal.com
}

\maketitle

%%%%%%%%% ABSTRACT
\begin{abstract}
This paper presents a method for gaze estimation according to face images.
We train several gaze estimators adopting four different network architectures, including an architecture designed for gaze estimation (i.e.,iTracker-MHSA) and three originally designed for general computer vision tasks(i.e., BoTNet, HRNet,  ResNeSt). 
Then, we select the best six estimators and ensemble their predictions through a linear combination.
The method ranks the first on the leader-board of ETH-XGaze Competition, achieving an average angular error of $3.11^{\circ}$ on the ETH-XGaze test set.% 补一下官方的比赛名称
\end{abstract}

%%%%%%%%% BODY TEXT
\section{Introduction}

Estimating people's eye-gaze according to facial images plays a fundamental role in varied applications of  human-computer interaction\cite{jacob2003eye,morimoto2005eye,majaranta2014eye}, affective
computing\cite{d2012gaze}, and medical diagnosis\cite{wang2015atypical,jiang2017learning}. Though the
gaze estimation task can be efficiently solved with
deep learning-based approaches, we lack an appropriate and unified metric to compare different state-of-the-art methods. Additionally, existing gaze estimation datasets have limitations on the head pose and gaze variations and imperfect quality of images and ground truth labels. To advance the development of gaze estimation research,  Zhang\cite{zhang2020eth} proposed ETH-XGaze, a new gaze estimation dataset  including more than one million high-resolution images of varying gaze and head poses.

Considering a larger variety of settings in ETH-XGaze, including variation of viewpoint, extreme gaze angles, lighting variation, and occluders like glasses,
it is a challenge to conduct  gaze estimation accurately on it. Basic neural models like Resnet-50 have limitation on providing accurate enough in a
varying environment, and more delicate network architectures need to be explored for higher accuracy.

In this paper, we propose a gaze estimation network based on iTracker\cite{GazeCapture} and three other architectures for general computer vision purposes. We explore the utility of muti-scale, split-attention networks, and different training techniques in gaze estimation tasks.

\section{Relate Work}
%There has been  plenty of work on predicting gaze.
%Here, we give a brief overview of some of the existing gaze estimation methods. Then, we review the methods about network architecture design related to our work.
%\subsection{Gaze Estimation}
Gaze estimation methods can be classified into model-based and appearance-based methods. Model-based methods generally use a geometric eye model with parameters inferred from localized eye landmarks such as the pupil center\cite{guestrin2006general} and the iris edge\cite{yamazoe2008remote,chen20083d} to obtain gaze estimation result. Model-based methods usually require the specific device for high- resolution images and are limited by light conditions and short working distance, which is not suitable for gaze estimation under varying environments. Meanwhile, appearance-based methods directly estimate the gaze direction from the face images.  Appearance-based methods\cite{GazeCapture,zhang2017mpiigaze,cheng2018appearance} mostly learn a mapping from face or eyes images to gaze and rely on deep learning to extract feature from images. Appearance-based methods can estimate gazes despite light or appearance variations with the help of enough training data. 

Based on previous works, different methods for learning based gaze estimation have been proposed. Cheng\cite{chen2018appearance} proposed to use dilated convolutions and Cheng\cite{cheng2020coarse} came up with a coarse-to-fine network to improve gaze estimation result. Chen and Zhang \cite{cheng2020gaze} proposed to improve gaze estimation by exploring two-eye asymmetry.

%\subsection{Backbone}
%% 这一段可以不写了
%-------------------------------------------------------------------------
\section{Methods}
The gaze estimation task is formulated as a regression from the normalized face images to the pitch-yaw gaze direction vectors. 
In ETH-XGaze Challenge, we adopted four different network architectures to train the gaze estimators, including an architecture designed for gaze estimation (i.e., iTracker-MHSA) and three originally designed for general computer vision tasks (i.e., BoTNet, HRNet, ResNeSt). 
Then, we select the best six estimators and ensemble their predictions through a linear combination. 
Below, we introduce details of the adopted four network architectures.
%The core of obtaining accurate gaze estimation results is to extract gaze-related features from face effectively. Here we introduce several gaze estimation network architectures which can obtain valuable information beneficially from the full face for gaze estimation.
\begin{figure*}[t]
\begin{center}
\includegraphics[width=.9\linewidth]{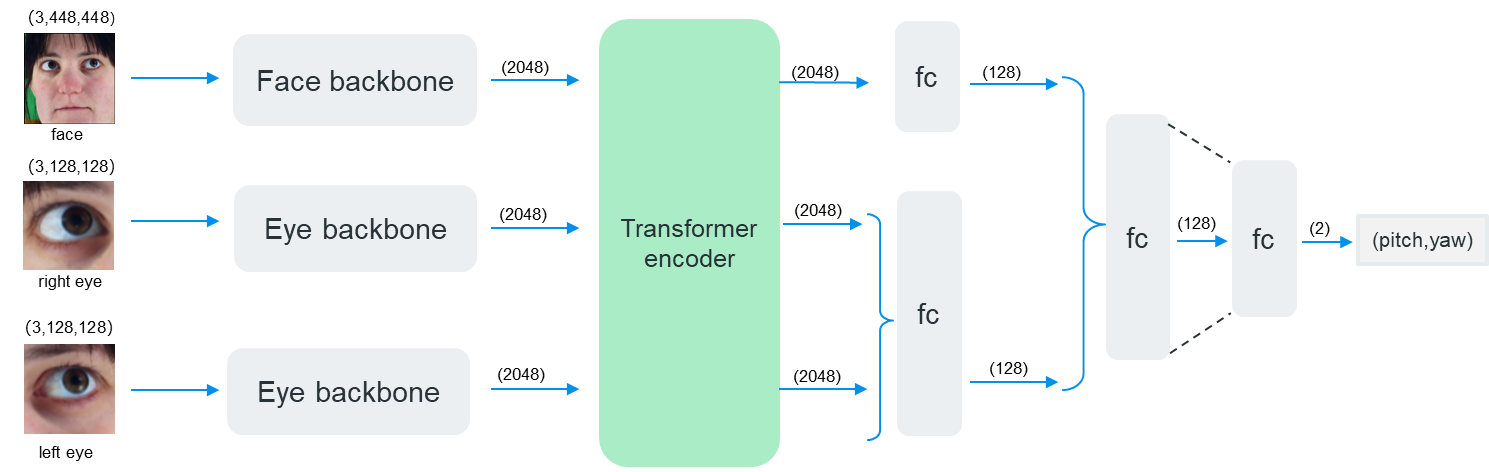}
\end{center}
   \caption{An overview of iTracker-MHSA architecture. The face backbone is Resnet50 and the eye backbone is dilated Resnet50. Shapes of features is labelled in round brackets.}
\label{fig:iTracker}
\end{figure*}

\begin{figure}[t]
\begin{center}
\includegraphics[width=.9\linewidth]{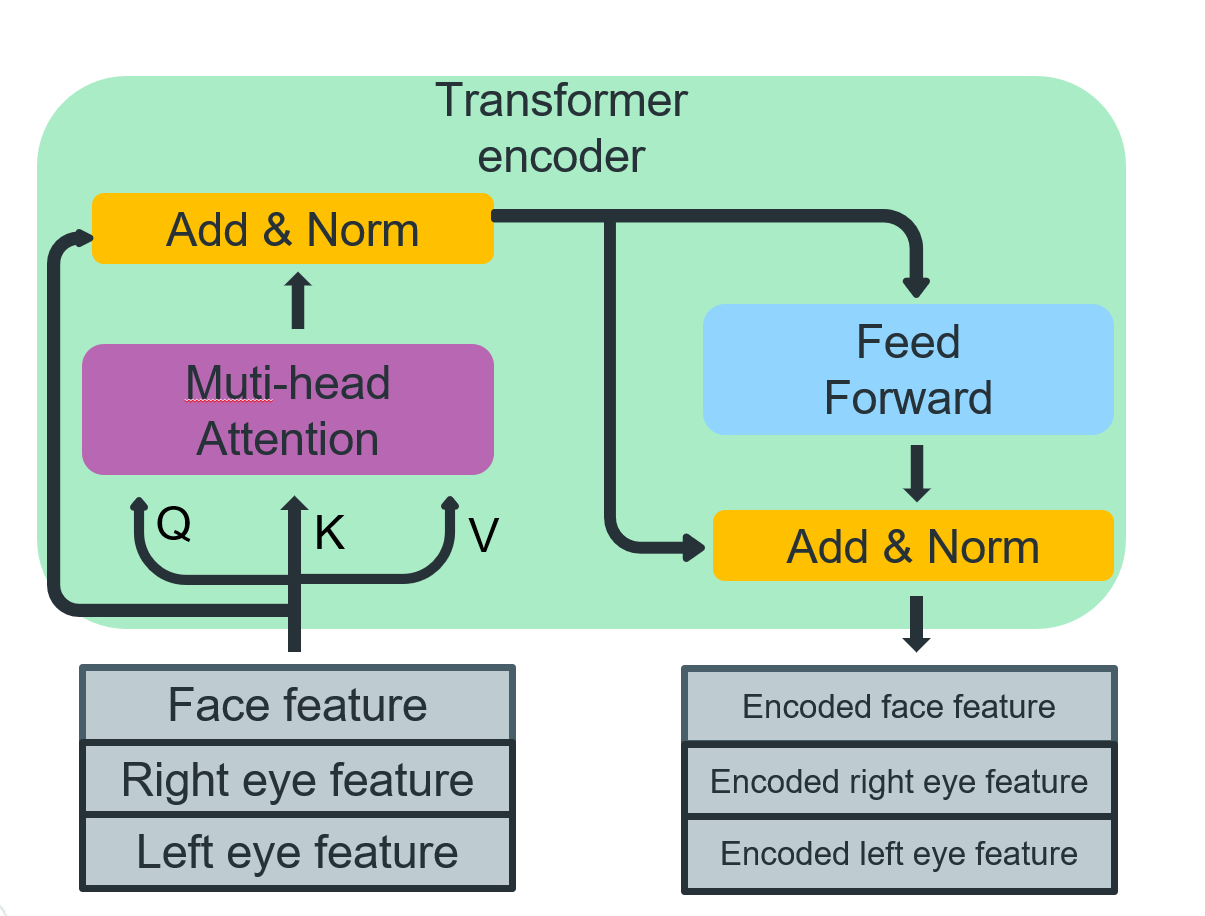}
\end{center}
   \caption{The  structure of transformer encoder.}
\label{fig:transformer}
\end{figure}
\subsection{iTracker-MHSA}
To solve gaze point estimation problem, Krafka et al. \cite{GazeCapture} proposed iTracker, which combines the information from left and right eye images,
face images along with face grid information. 
The face grid indicates the position of the face region in the captured image for gaze point estimation.
We make some improvements on iTracker for our gaze direction estimation task and the key modifications in terms of input and architecture compared to iTracker are summarized as follows.

Figure.~\ref{fig:iTracker} illustrates the architecture of iTracker-MHSA.
Firstly we remove the face grid branch since our gaze direction labels have been normalized in a norm camera space.
Then we substitute eye backbone of iTracker for dilated Resnet50\cite{he2016deep} and face backbone for convolution layers of Resnet50 to obtain better features. 
According to Chen\cite{cheng2020gaze}, compared with canonical convolutions, dilated convolutions achieve remarkable accuracy gains on gaze estimation tasks.
Last, to learn the relationship of face and eye features and adjust weights of face and eye features automatically, we use a multi-head self-attention mechanism in our designed network iTracker-MHSA. Specifically, we add a transformer encoder\cite{vaswani2017attention} to encode face and eye features, and then we use encoded features to estimate gaze. 

The sub-network structure of the transformer encoder is shown in Fig.\ref{fig:transformer}. A transformer encoder has two sub-layers. The first is a multi-head self-attention layer, and the second is a fully connected feed-forward network. We implement a residual connection around each of the two sub-layers and use layer normalization to normalize the sum.
As a result, the output of each sub-layer is LayerNorm(x + Sublayer(x)), where Sublayer is a multi-head self-attention layer or feed-forward layer. The detail of the multi-head self-attention mechanism can be found in \cite{vaswani2017attention}. Given face and eye features extracted by backbone as input, the encoder output encoded features with the same shape of input.

To locate eye positions in given face images, we use hrnet\_w18\cite{wang2020deep} to detect facial landmarks. Since detected facial landmarks are decimals, we use RoI align \cite{he2017mask} to crop eye images for accurate eye location.

During training iTracker-MHSA, online hard example mining strategy \cite{shrivastava2016training} are conducted to ensure models' ability to handle hard examples. Specifically, we sort losses of samples in a batch in descending order and double the losses of the top 30\% samples.

%\subsection{Architec.tures for general computer vision tasks}
 
\subsection{BoTNet}
A lot of network architectures have been proposed to extract images feature and improve downstream tasks such as image classification, object detection, and so on. Some of them are proved to be effective in our gaze estimation task.
    
\begin{figure}[t]
\begin{center}
\includegraphics[width=.9\linewidth]{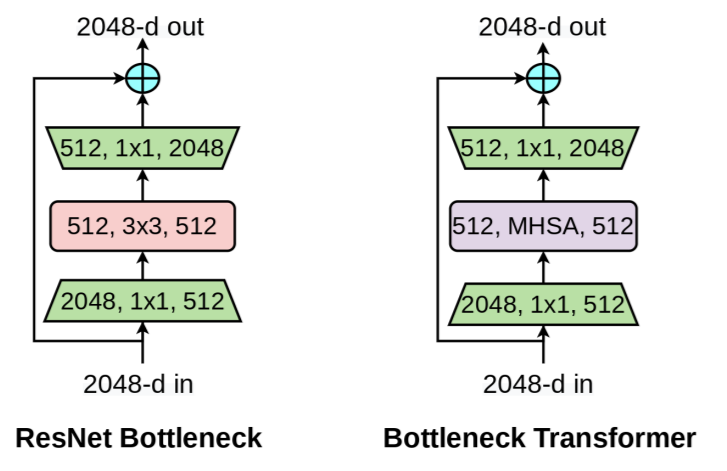}
\end{center}
\caption{Left: A ResNet Bottleneck Block, Right: A Bottleneck Transformer block. The difference is the replacement of the spatial 3 × 3 convolution layer with
Multi-Head Self-Attention.\cite{srinivas2021bottleneck}}
\label{fig:botnet}
\end{figure}

\begin{figure}[t]
\begin{center}
\includegraphics[width=.7\linewidth]{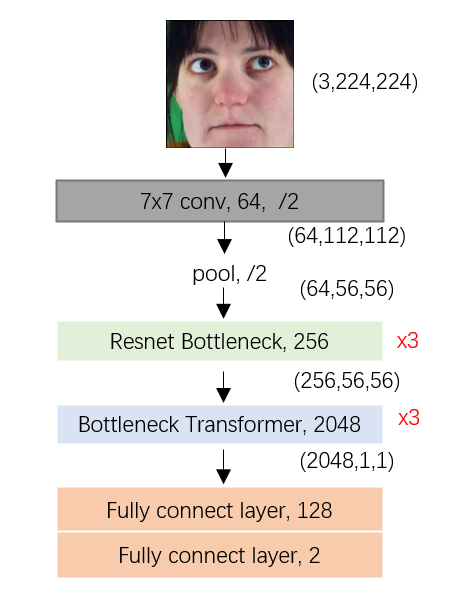}
\end{center}
   \caption{An overview of our boTNet based network. The shapes of features are labelled in round brackets.}
\label{fig:botnet2}
\end{figure}

BoTNet\cite{srinivas2021bottleneck} proposed a method to replace spatial convolution layers with the multi-head self-attention layer proposed in the Transformer\cite{vaswani2017attention}, which helps network learning global features of the input. 
Fig.\ref{fig:botnet} shows how to change Resnet bottleneck block to a bottleneck transformer block. 
The structure of the multi-head self-attention layer is described in \cite{srinivas2021bottleneck}.
We design a network based on BoTNet as shown in Fig.\ref{fig:botnet2}. 
We down-sample the input feature map using convolution with stride 2 and max pooling and then obtain a 2048-dimensional feature by three resnet bottleneck blocks and three bottleneck transformer blocks, followed by two fully connected layers to finish gaze estimation.

\subsection{HRNet}
High-resolution representation learning for eye plays an important role in gaze estimation. Different from other networks which downsampling feature maps, HRNet\cite{sun2019high} develops a creative method to maintain the high-resolution representation of input through the whole inference process of model. HRNet connects high-to-low resolution convolutions in parallel and repeatedly implements fusions across parallel convolutions for high-resolution representations.

We conduct HRNetV2-W64\cite{sun2019high} for learning a 1000-dim high-resolution representation of input face images. Based HRNet backbone, we add two fully connection layers (1000-128-2) to learning a mapping from the 1000-dim representation to the 2-dim gaze result.

\subsection{ResNeSt}
ResNeSt\cite{zhang2020resnest} is a network architecture that combines channel-wise attention with multi-path representation strategies to improve the network representation ability. A block of ResNeSt
perfroms a series of transformations on low dimensional features and concatenates their outputs as in a multi-path network. Each transformation conducts a channel-wise attention strategy to capture the relationship of different feature maps.

Like HRNet method, we use ResNeSt269\cite{zhang2020resnest} as our backbone to extract features and employ three fully connection layers (2048-128-128-2) following the backbone to estimate gaze direction.

\section{Experiments}
We train the gaze estimators on ETH-XGaze dataset using the above four architectures with different settings. 
Then we choose six estimators that performed the best and ensemble their predictions to get the final prediction.
%We implement methods mentioned before on ETH-XGaze dataset and here are the details of the experiment settings, process, and results.
\subsection{Data pre-processing}
\textbf{Image size.} We directly used the challenge-provided 224$\times$224 facial images which are normalized by \cite{zhang2018revisiting}. We resize the input frame to different sizes (e.g., 448$\times$448, 640 $\times$ 640)  by bilinear interpolation when training and test models, and experiments show changing image size is vital to gaze estimation.

\textbf{Data augmentation.} We flip training images horizontally to augment data. The yaw labels of flipped data are opposites of original labels and the pitch labels are changeless.

\begin{figure}[t]
\begin{center}
\includegraphics[width=1\linewidth]{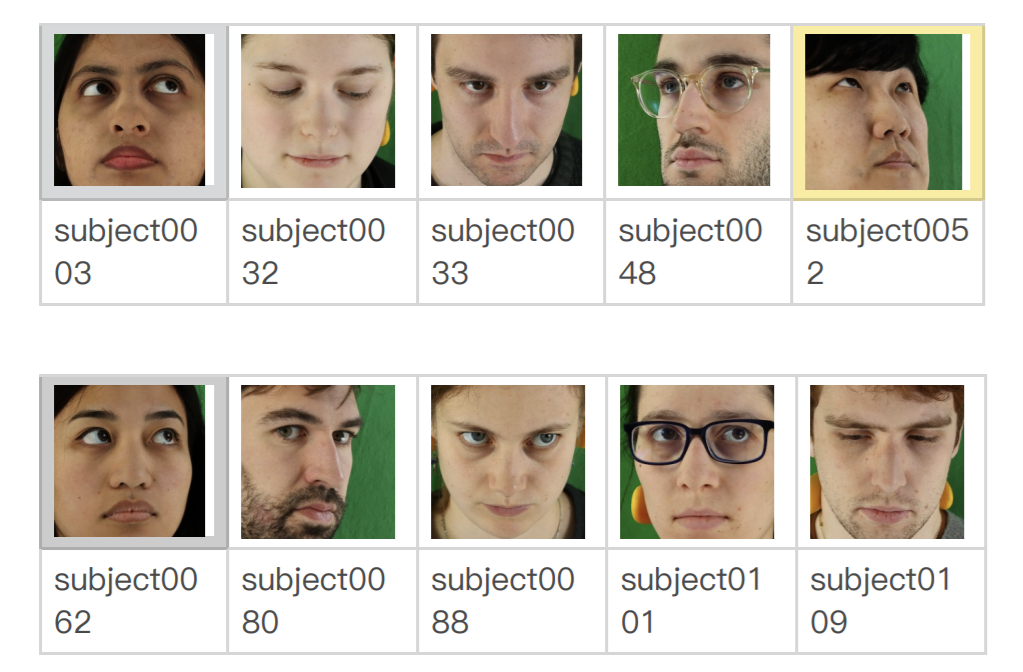}
\end{center}
   \caption{Examples of our validation set.}
\label{fig:validation}
\end{figure}

\textbf{Validation set split.}
We choose 10 subjects as a validation set from the origin training set by sex, skin tone, and whether to wear glasses. The subject ids of validation set are: 03, 32, 33, 48, 52, 62, 80, 88, 101, 109 and some examples are shown in Fig \ref{fig:validation}.
We train all models only using training set and choose best models on the
validation set to conduct inference on test set in our experiments.
\subsection{Implementation details}
\ We train the models on  8 NVIDIA V100  GPUs using DistributedDataParallel(DDP) Pyorch and synchronized batch normalization strategy.
All models we use are pretrained on Imagenet\cite{5206848}.
We use L1 loss function, exponential learning rate schedule strategy and adam\cite{kingma2014adam} optimizer to train models.

For \textbf{iTracker-MHSA} with input size 448, the batch size is  (30 $\times$ 8),  the initial learning rate is (1e-4 $\times$ 8) .
We choose the best performed model on the validation set among different training epochs. % 在哪些集合里选最佳模型。比如，在不同epoch的模型堆里挑一个最佳的。

For \textbf{BoTNet} with input size 224, the batch size is  (24 $\times$ 8) and the initial learning rate is (1e-4 $\times$ 8).

For \textbf{ResNeST} with input size 448, the batch size is  (12 $\times$ 8),  the initial learning rate is (1e-4 $\times$ 8), and the chosen epoch is 10.

As for \textbf{HRNet}, three models are selected. We use horizontal flip to augment the training set and step learning rate schedule strategy when training them.  

For HRNet with input size 640, the batch size is  (24 $\times$ 8) and  the initial learning rate is (2.5e-5 $\times$ 8). And for HRNet with input size 768, the batch size is  (16 $\times$ 8) and the initial learning rate is (2.5e-5 $\times$ 8). For HRNet with input size 896, the batch size is  (12 $\times$ 8) and the initial learning rate is (2.5e-5 $\times$ 8).

\subsection{Experiment results}
\subsubsection{Single model results}
We train different models in different input size settings and report results on the test set in Table \ref{table:single}.
\begin{table}[]
\begin{tabular}{|c|c|c|}
\hline
method         & input size & gaze error \\ \hline
iTracker       & 224        & 4.02       \\
iTracker with dilated CNN      & 224        & 3.80      \\
iTracker\_MHSA & 224        & 3.54       \\
iTracker\_MHSA & 448        & 3.37       \\ \hline
BoTNet         & 224        & 3.84       \\ \hline
HRNet          & 224        & 3.84       \\
HRNet          & 448        & 3.50       \\
HRNet          & 640        & \textbf{3.22}      \\
HRNet          & 768        & 3.37       \\
HRNet          & 896        & 3.39       \\ \hline
ResNeSt        & 224        & 3.84       \\
ResNeSt        & 448        & 3.34       \\ \hline
\end{tabular}
\caption{Averaged angular errors of single models}
\label{table:single}
\end{table}

\subsubsection{Ensembled model results}
Experiments prove the weighted average of different single-model results is better than the best single model result in Table \ref{table:single}.Our ensembled model results is shown in Table \ref{table:Ensemble1} and Table \ref{table:Ensemble2}. The weight of different models is chosen empirically and the lowest gaze error of our final result is 3.11.

\begin{table}[]
\begin{tabular}{|c|c|c|c|}
\hline
method         & input size & weight & gaze error \\ \hline
iTracker\_MHSA & 448       & 0.33  & 3.42        \\ 
HRNet          & 640      & 0.33  & 3.22     \\
ResNeSt        & 448       & 0.33  & 3.34      \\ \hline
average of above &  /   & / &  \textbf{3.14} \\ \hline
\end{tabular}

\caption{Average angular errors of the ensemble results of 3 models}
\label{table:Ensemble1}
\end{table}

\begin{table}[]
\begin{tabular}{|c|c|c|c|}
\hline
method         & input size & weight & gaze error \\ \hline
iTracker\_MHSA & 448       & 0.2  & 3.42        \\ 
BoTNet          & 224      & 0.1  & 3.84     \\
HRNet          & 640      & 0.4  & 3.22     \\
HRNet          & 768      & 0.1  & 3.37     \\
HRNet          & 896      & 0.1  & 3.39     \\
ResneSt        & 448       & 0.1  & 3.34      \\ \hline
weighted average &  /   & / &  \textbf{3.11} \\ \hline
\end{tabular}

\caption{Average angular errors of the ensemble results of 6 models.}
\label{table:Ensemble2}
\end{table}

{\small
\bibliographystyle{ieee_fullname}
\bibliography{egbib}
}

\end{document}